\documentclass[default, iicol]{sn-jnl}
\usepackage{graphicx}%
\usepackage{booktabs}
\usepackage{amsmath,amssymb,amsfonts}%
\usepackage{amsthm}%
\usepackage{array}
\usepackage{longtable}
\usepackage{caption}
\usepackage{mathrsfs}%
\usepackage[title]{appendix}%
\usepackage{xcolor}%
\usepackage{textcomp}%
\usepackage{manyfoot}%
\usepackage{stfloats}
\usepackage{booktabs}%
\usepackage{algorithm}%
\usepackage{multirow}
\usepackage{algorithmicx}%
\usepackage{algpseudocode}%
\usepackage{listings}%
\usepackage{flushend}

\begin{document}

\title[Article Title]{Towards Explainable AI: Multi-Modal Transformer for Video-based Image Description Generation}

\author[1]{\fnm{Lakshita} \sur{Agarwal}}\email{lakshitaa3@gmail.com}
\author*[2]{\fnm{Bindu} \sur{Verma}}\email{bindu.cvision@gmail.com}
\affil[1]{\orgdiv{Department of Information Technology}, \orgname{Delhi Technological University}, \orgaddress{\city{Delhi}, \postcode{110042}, \country{India}}}
\affil*[2]{\orgdiv{Assistant Professor, Department of Information Technology}, \orgname{Delhi Technological University}, \orgaddress{\city{Delhi}, \postcode{110042}, \country{India}}}

\abstract{Understanding and analyzing video actions are essential for producing insightful and contextualized descriptions, especially for video-based applications like intelligent monitoring and autonomous systems. The proposed work introduces a novel framework for generating natural language descriptions from video datasets by combining textual and visual modalities. The suggested architecture makes use of ResNet50 to extract visual features from video frames that are taken from the Microsoft Research Video Description Corpus (MSVD), and Berkeley DeepDrive eXplanation (BDD-X) datasets. The extracted visual characteristics are converted into patch embeddings and then run through an encoder-decoder model based on Generative Pre-trained Transformer-2 (GPT-2). In order to align textual and visual representations and guarantee high-quality description production, the system uses multi-head self-attention and cross-attention techniques. The model's efficacy is demonstrated by performance evaluation using BLEU (1-4), CIDEr, METEOR, and ROUGE-L. The suggested framework outperforms traditional methods with BLEU-4 scores of 0.755 (BDD-X) and 0.778 (MSVD), CIDEr scores of 1.235 (BDD-X) and 1.315 (MSVD), METEOR scores of 0.312 (BDD-X) and 0.329 (MSVD), and ROUGE-L scores of 0.782 (BDD-X) and 0.795 (MSVD). By producing human-like, contextually relevant descriptions, strengthening interpretability, and improving real-world applications, this research advances explainable AI.}


\maketitle

\section{Introduction}
In computer vision and natural language processing, creating informative textual descriptions of events that occur in video sequences is a fundamental problem known as "video description generation"~\cite{aafaq2019video}. In contrast to static image captioning, video-based description generation involves an awareness of object interactions, temporal dependencies, and contextual linkages in order to provide narratives that are both contextually rich and coherent. Applications like intelligent surveillance, multimedia retrieval, autonomous driving, and human-computer interaction greatly benefit from these capabilities. Capturing dynamic scene changes, identifying pertinent objects, and producing descriptions that appropriately reflect the changing context are the challenges~\cite{li2025generative}~\cite{agarwal2024methods}. Conventional methods for comprehending videos, such as object detection and action recognition, mostly concentrate on figuring out "what" is happening in a scene. Explaining "why" things happen is more difficult, though, particularly in complex, multi-agent systems where behaviour is influenced by environmental suggestions~\cite{shoman2024enhancing}. Known as video captioning or video-based description generation, the process combines textual and visual modalities to generate interpretable and informative natural language descriptions~\cite{li2021caption}~\cite{gou2022driver}. The ability to produce context-aware and semantically relevant video descriptions has greatly improved with recent developments in deep learning, especially transformer-based systems.

This work presents a novel transformer-based architecture for generating video descriptions by effectively combining text and image modalities. Using a GPT-2 encoder-decoder with multi-head attention and ResNet50 for visual feature extraction, the model is trained on diverse datasets, including MSVD and BDD-X. Training is optimized with gradient accumulation and mixed precision to boost efficiency. Evaluated using BLEU (1–4), CIDEr, METEOR, and ROUGE-L, the model produces fluent, contextually relevant descriptions aligned with human annotations, contributing to advancements in video explainability and AI-driven understanding.

The major contributions of the proposed work are summarized below:
\begin{itemize}
    \item To provide natural language descriptions of video sequences, the following work proposes a transformer-based architecture that integrates a GPT-2-based language model with the visual features retrieved by ResNet50.    
    \item The method combines video datasets from various sources (MSVD, and BDD-X), allowing the model to generalise across domains. High-quality training samples are guaranteed via sophisticated preprocessing.
    \item Gradient accumulation and mixed precision training are used to optimise the model, increasing computing efficiency without sacrificing output quality.
    \item Fluency, contextual relevance, and coherence are ensured by thoroughly evaluating the generated descriptions using BLEU (1–4), CIDEr, METEOR, and ROUGE-L standards.
    \item Although this method is extremely beneficial for intelligent transportation and autonomous driving, it can also be used in other fields, including robotics, assistive technology, and surveillance.
\end{itemize}

The remaining sections of the paper are organized as follows: Section~\ref{relatedwork} examines relevant studies in multimodal learning and video description generation. The suggested methodology, including feature extraction, model architecture, and dataset preprocessing, is described in Section~\ref{methods}. Experimental results and assessment results are presented in Section~\ref{results}. Lastly, conclusions and possible future study directions are covered in Section~\ref{conclusion}.

\section{Related Work}
\label{relatedwork}
In computer vision and natural language processing, video-based description creation is an essential task that allows systems to produce textual summaries of visual content. The creation of video descriptions must consider temporal relationships, object interactions, and dynamic scene changes across frames, in contrast to static image captioning. Early video description generation approaches included recurrent neural networks (RNNs) or long short-term memory (LSTM) networks for text synthesis and convolutional neural networks (CNNs) for feature extraction. 

A Sequence-to-Sequence (Seq2Seq) model trained on MSVD~\cite{chen2011collecting} was presented by Venugopalan et al.~\cite{venugopalan2015sequence} to produce video descriptions. However, because of the limits of LSTMs, these models have trouble handling contextual inconsistencies and long-range dependencies. Later, to make generated descriptions more relevant, attention techniques were added~\cite{yao2015describing}. Transformer-based models have led to a considerable improvement in video captioning systems. End-to-end dense Video Captioning was proposed by Zhou et al.~\cite{zhou2018end}, who used spatiotemporal attention mechanisms to grasp multiple frames. Compared to conventional RNN-based models, MART (Memory-Augmented Recurrent Transformer), which was introduced by Lei et al.~\cite{lei2020mart}, performs more effectively at capturing long-term video dependencies. More recently, multimodal pretraining for video-language understanding has been investigated by VideoBERT~\cite{sun2019videobert} and ClipBERT~\cite{lei2021less}. These methods achieve state-of-the-art outcomes on datasets such as MSVD~\cite{chen2011collecting} and MSR-VTT~\cite{xu2016msr} by aligning textual and visual representations. Datasets such as Berkeley DeepDrive eXplanation (BDD-X)~\cite{kim2018textual} have been useful for producing explainable AI-driven annotations for vehicle-based video description generation. Studies like Shoman et al.~\cite{shoman2024enhancing} integrated explainability processes into vision-language models and concentrated on justification-based video descriptions for autonomous driving. An attention-guided transformer was developed by Cui et al.~\cite{cui2024survey} to improve contextual reasoning in captioning models based on BDD-X.

Recent advancements in multimodal learning combine text, audio, and visual data. Hori et al.\cite{hori2017attention} introduced an audio-visual attention model using speech and scene context, while Yu et al.\cite{yu2019multimodal} enhanced captioning on datasets like Flickr8k and MSVD using spatiotemporal object graphs. These methods improve contextual awareness, particularly in dynamic scenarios like vehicle interactions. Despite progress, generating accurate, human-like video descriptions remains challenging. The proposed approach addresses this by combining a GPT-2 encoder-decoder with ResNet50-based feature extraction, outperforming prior methods on BLEU (1–4), CIDEr, METEOR, and ROUGE-L. It refines multimodal embeddings for context-aware captions and boosts efficiency with gradient accumulation and mixed precision training. By bridging explainable vehicular (BDD-X) and general (MSVD) video captioning, the model delivers interpretable, context-rich descriptions suitable for real-world use.

\section{Overview of the Proposed Work}
\label{methods}
\begin{figure*}[!h]
\centering
\includegraphics[width=1\textwidth]{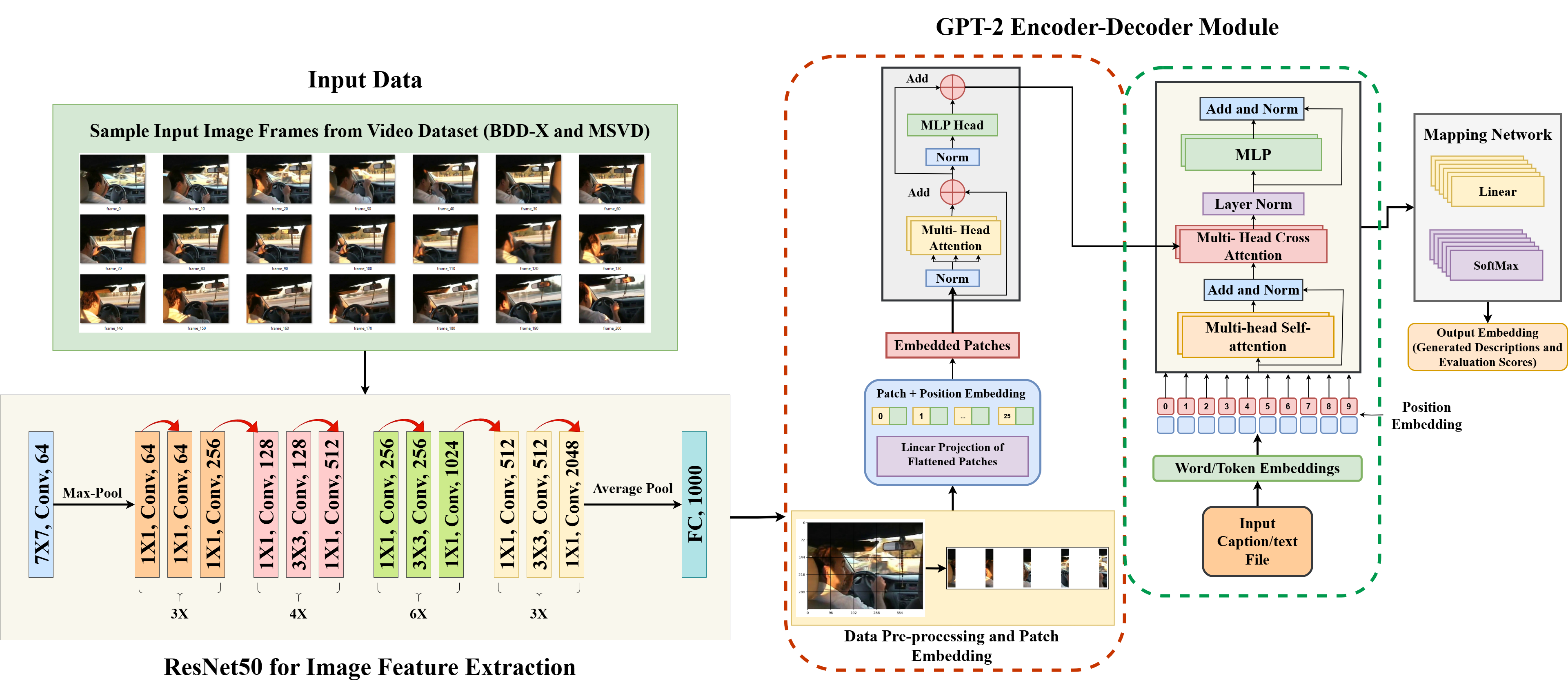}
\caption{Framework for the Proposed Model (ResNet50-GPT2): The system incorporates ResNet50 for image feature extraction and a GPT-2 encoder-decoder model to generate context-aware video-based image descriptions.}
\label{architecture}
\end{figure*}

The proposed study introduces a transformer-based, context-aware reasoning framework for generating logical and relevant video descriptions. Leveraging the BDD-X and MSVD datasets, the system learns from both action-justified driving videos and diverse scene representations. It employs an optimized GPT-2 encoder-decoder model with multi-head attention and ResNet50-based visual feature extraction to enhance contextual understanding. Training is guided by action-justification pairs from BDD-X and captioned sequences from MSVD, promoting coherent, human-like output. Gradient accumulation and mixed precision training boost computational efficiency without compromising accuracy. Evaluations using BLEU (1–4), CIDEr, METEOR, and ROUGE-L show superior performance over traditional methods. The basic architecture of the proposed framework is shown in Figure~\ref{architecture}, which further demonstrates the pipeline for data pre-processing, model training, and description generation.

\subsection{Data Pre-processing}
The input data includes video frames from the BDD-X and MSVD datasets. Video frames are extracted, pertinent images are filtered, and then the frames are transformed into structured input for the model as part of the pre-processing pipeline. ResNet50, which converts spatial information from images into high-dimensional feature vectors, is used for feature extraction. After that, a linear projection and position encoding method are used to patch and embed these feature vectors. The GPT-2 encoder-decoder module then receives the processed embeddings and aligns the word/token embeddings with the appropriate visual features. To further improve model resilience, tokenization, data augmentation, and train-test splits are used.

\subsection{Model Architecture}
The proposed method combines a GPT-2 encoder-decoder model to generate context-aware descriptions of vehicle actions with ResNet50 to extract image features. The model is fine-tuned on the BDD-X and MSVD datasets to become optimal in dynamic scenarios for generating meaningful descriptions. The training process employs gradient accumulation and mixed precision training to achieve optimal computing efficiency. 

ResNet50 acts as a feature extractor by passing each input frame \( x \in \mathbb{R}^{3 \times H \times W} \) through a series of convolutional, batch normalization, ReLU, and residual layers. The output feature vector \( f \) is computed as:
\begin{equation}
f = \text{ResNet50}(x) = \text{AvgPool}(F_{\text{res}}(x))
\end{equation}
where, \( F_{\text{res}} \) denotes the output of the final convolutional block, and \( \text{AvgPool} \) is the global average pooling operation applied to extract the final feature representation \( f \in \mathbb{R}^{2048} \).

Using position embeddings \( P \in \mathbb{R}^{N \times d} \) and a learnable linear projection \( W_p \in \mathbb{R}^{2048 \times d} \), the visual characteristics retrieved by ResNet50 are transformed into embedded visual tokens:
\begin{equation}
E_v = f W_p + P
\end{equation}

The GPT-2 model uses these embedded patches as input tokens to produce descriptions that are logical and sensitive to context. The model utilizes self-attention mechanisms to enhance contextual understanding, formulated as:
\begin{equation}
\text{Self-Attention}(Q, K, V) = \text{Softmax}\left(\frac{QK^T}{\sqrt{d_k}}\right) V
\end{equation}
where \( d_k \) is the key vector dimension, and \( Q, K, V \) represent the query, key, and value matrices.

To maintain the semantic relevance of the generated descriptions to video-based activities, the multi-head cross-attention method aligns image embeddings with textual descriptions. The decoder uses attention-weighted contextual embeddings to improve the textual output:
\begin{equation}
\text{Decoder-Attention}(Q, K, V) = \text{Softmax}\left(\frac{QK^T}{\sqrt{d_k}}\right) V
\end{equation}

The final word prediction probability is computed as:
\begin{equation}
y_t = \text{Softmax}(W_o h_t + b_o)
\end{equation}
where \( W_o \) and \( b_o \) are the output weight matrix and bias, respectively, and \( h_t \) represents the hidden state at time step \( t \).

To enhance generalization, the model employs a combined loss function incorporating cross-entropy loss and L2 regularization:
\begin{equation}
\mathcal{L} = - \sum_{t=1}^{T} \log P(y_t \mid y_{<t}) + \lambda \| \theta \|^2
\end{equation}
where \( \lambda \) is the regularization parameter and \( P(y_t \mid y_{<t}) \) denotes the probability of predicting \( y_t \) given prior words.

In BLEU-4, CIDEr, METEOR, and ROUGE-L metrics, the proposed framework outperforms traditional methods, ensuring that the generated descriptions are sensible, logical, and context-aware. These metrics are utilized to analyze the contextual validity and linguistic quality of the descriptions generated by the GPT-2 model for unseen videos upon training. The method achieves performance improvement over traditional methods by contrasting the generated descriptions with ground-truth annotations. Ultimately, the framework generates natural language narratives that effectively describe dynamic visual scenes, promoting explainability in a range of applications.

The following summarizes Algorithm~\ref{algo} for the proposed model and its evaluation:
\begin{algorithm}[!h]
\caption{Context-Aware Description Generation}
\label{algo}
\begin{algorithmic}[1]
\State \textbf{Input:} Video frames (BDD-X, MSVD), image-caption pairs (Flickr8k)
\State \textbf{Output:} Generated descriptions, Evaluation Scores

\State \textbf{1. Preprocessing and Embedding}
\State Extract and embed image frames using patch encoding
\State Tokenize and format textual descriptions

\State \textbf{2. Dataset Split}
\State Create balanced train-test sets with captions and action-justification pairs

\State \textbf{3. Feature Extraction}
\State $\text{Visual\_features} \gets \text{ResNet50}(I)$
\State Project features and tokenize text using GPT-2

\State \textbf{4. Model Training}
\State Initialize GPT-2 with multi-head attention and cross-modal fusion
\State Train with AdamW, gradient accumulation, and mixed precision

\State \textbf{5. Inference and Evaluation}
\For{each test sample}
    \State Generate and decode descriptions
\EndFor
\State Compute BLEU-4, CIDEr, METEOR, ROUGE-L

\State \textbf{6. Output}
\State Present evaluation scores and visualize results
\end{algorithmic}
\end{algorithm}

\section{Experimental Analysis}
\label{results}
The proposed model was implemented in PyTorch and evaluated on Google Colab Pro+ using 52\,GB RAM, an NVIDIA A100 GPU (40\,GB VRAM), and a virtual Intel Xeon CPU. It generates context-aware descriptions using a GPT-2 encoder-decoder (approximately 66.7M parameters) and ResNet50 for visual feature extraction. Input frames were resized to \texttt{(batch\_size, 224, 224, 3)}. Training was conducted over 75 epochs with the AdamW optimizer (learning rate $1 \times 10^{-5}$, weight decay 0.01), using a batch size of 32 and gradient accumulation. Mixed precision training and gradient clipping were applied to improve computational efficiency and training stability. The model was trained using the MSVD dataset\footnote{\url{https://www.kaggle.com/datasets/vtrnanh/msvd-dataset-corpus}}, which contains short descriptive captions for diverse video clips, and the BDD-X dataset\footnote{\url{https://github.com/JinkyuKimUCB/BDD-X-dataset}}, which includes over 26K annotated actions across 8.4M frames and 6,970 videos. During preprocessing, linguistic descriptions were tokenized and structured into action-justification pairs, while image frames were embedded via ResNet50 and projected linearly into patch embeddings. These embeddings were then fed into the multi-head self-attention module of the GPT-2 encoder-decoder to align image and text contexts effectively. Performance was evaluated on a 20\% test split using BLEU-4, CIDEr, METEOR, and ROUGE-L. The model consistently outperformed conventional captioning techniques, demonstrating its ability to generate logical, contextually accurate descriptions across diverse visual inputs. The architecture’s use of self- and cross-attention mechanisms ensured semantic accuracy and narrative coherence, reinforcing its applicability in real-world visual captioning tasks. The following Table~\ref{experiment} depicts the overall architectural structure of the proposed framework.
\begin{table*}[!htbp]
\centering
\caption{Experimental Setup and Performance Metrics}
\resizebox{0.60\textwidth}{!}{
\begin{tabular}{|l|l|}
\hline
\textbf{Component} & \textbf{Details} \\ \hline
\textbf{Framework Used} & PyTorch \\ \hline
\textbf{Image Feature Extractor} & ResNet-50 \\ \hline
\textbf{Model} & GPT-2 Transformer-based Encoder-Decoder \\ \hline
\textbf{Input Size} & (224, 224, 3) \\ \hline
\textbf{Batch size} & 32 \\ \hline
\textbf{Number of Epochs} & 75 \\ \hline
\textbf{Learning Rate} & \(1 \times 10^{-5}\) \\ \hline
\textbf{Optimizer} & AdamW \\ \hline
\textbf{Training Strategy} & Gradient Accumulation, Mixed-Precision Training \\ \hline
\textbf{Datasets Used} & BDD-X, MSVD, Filtered Flickr8k \\ \hline
\textbf{Evaluation Metrics} & BLEU 1-4, CIDEr, METEOR, ROUGE-L \\ \hline
\end{tabular}}
\label{experiment}
\end{table*}

\subsection{Ablation Study:} 
\begin{figure*}[!htbp]
    \centering
    \includegraphics[width=0.75\textwidth]{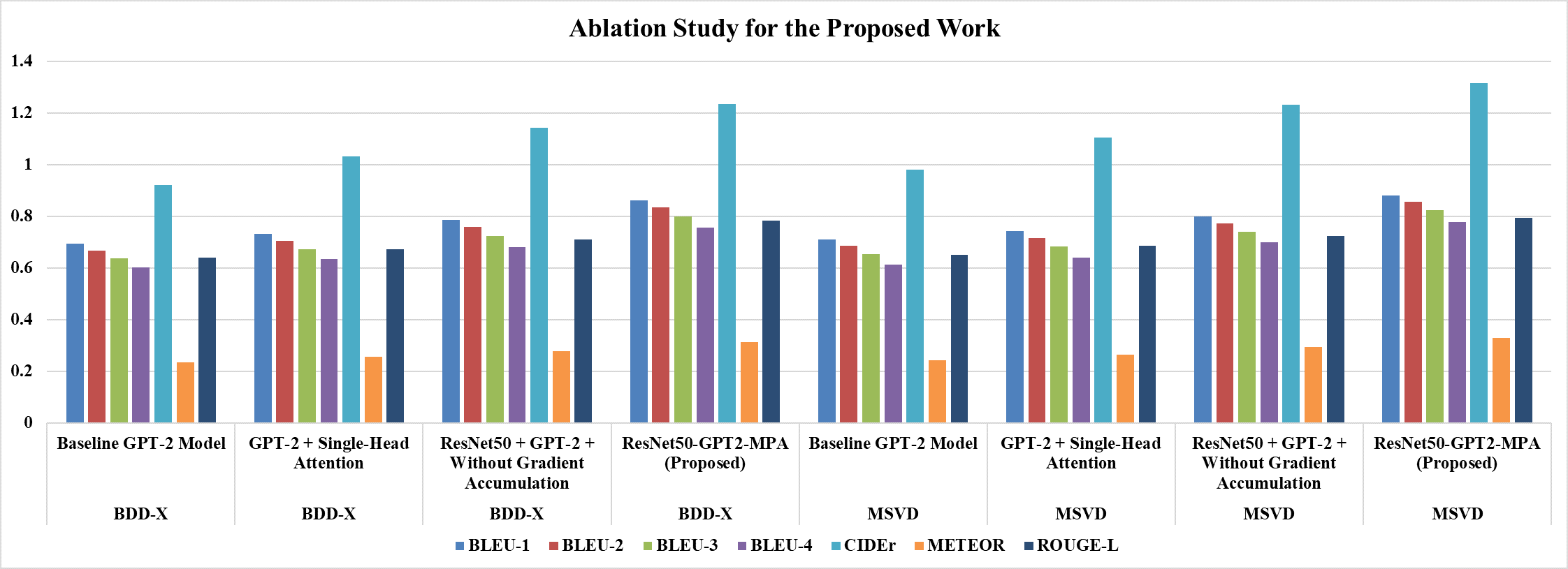}
    \caption{Demonstration of Ablation Study for the Proposed Work}
    \label{fig:ablation}
\end{figure*}

Utilising the BDD-X and the MSVD datasets, an ablation research was carried out to evaluate the contribution of various components in the suggested model. The research methodically eliminated or altered important architectural components, such as multi-head attention, gradient accumulation and mixed precision training optimisations, and visual feature extraction (ResNet-50), in order to assess each component's effect on model performance.  Model modifications were compared using the BLEU-1 to BLEU-4, CIDEr, METEOR, and ROUGE-L metrics, which provided a thorough assessment of the effects of each component on the quality of description production. As can be seen from Table~\ref{ablation}, the results demonstrate the importance of each element in producing textual descriptions for video frames that are both logical and contextually rich.
\begin{table*}[!htbp]
\centering
\caption{Ablation Study: Comparing the Impact of Model Components on Different Datasets}
\label{ablation}
\resizebox{\textwidth}{!}{
\begin{tabular}{|l|p{5cm}|c|c|c|c|c|c|c|}
\hline
\textbf{Dataset} & \textbf{Model Variant} & \textbf{BLEU-1} & \textbf{BLEU-2} & \textbf{BLEU-3} & \textbf{BLEU-4} & \textbf{CIDEr} & \textbf{METEOR} & \textbf{ROUGE-L} \\ 
\hline
\parbox[t]{1.5cm}{BDD-X}  
& Baseline GPT-2 Model & 0.693 & 0.668 & 0.638 & 0.601 & 0.921 & 0.234 & 0.640 \\ 
& GPT-2 + Single-Head Attention & 0.731 & 0.705 & 0.673 & 0.634 & 1.032 & 0.257 & 0.672 \\ 
& GPT-2 + Without Gradient Accumulation & 0.785 & 0.758 & 0.723 & 0.681 & 1.142 & 0.278 & 0.710 \\ 
& \textbf{ResNet50-GPT2 (Proposed)} & \textbf{0.860} & \textbf{0.835} & \textbf{0.800} & \textbf{0.755} & \textbf{1.235} & \textbf{0.312} & \textbf{0.782} \\ 
\hline
\hline
\parbox[t]{1.5cm}{MSVD}  
& Baseline GPT-2 Model & 0.710 & 0.685 & 0.654 & 0.612 & 0.980 & 0.242 & 0.651 \\  
& GPT-2 + Single-Head Attention & 0.742 & 0.715 & 0.683 & 0.641 & 1.105 & 0.265 & 0.685 \\  
& GPT-2 + Without Gradient Accumulation & 0.798 & 0.772 & 0.740 & 0.698 & 1.231 & 0.293 & 0.724 \\  
& \textbf{ResNet50-GPT2 (Proposed)} & \textbf{0.880} & \textbf{0.855} & \textbf{0.823} & \textbf{0.778} & \textbf{1.315} & \textbf{0.329} & \textbf{0.795} \\  
\hline
\end{tabular}}
\end{table*}
The model's performance on the MSVD and BDD-X datasets highlights the importance of each architectural component. The most significant impact came from removing ResNet-50, leading to major performance drops—BLEU-4 and CIDEr declined by 15.4\% and 25.4\% on BDD-X, and 21.3\% and 25.5\% on MSVD—emphasizing the need for strong visual features. Replacing multi-head attention with a single-head variant caused moderate losses, notably a 16.0\% drop in CIDEr on MSVD, showing its role in capturing fine-grained text-visual interactions. Omitting gradient accumulation reduced training efficiency and led to average BLEU-4 and CIDEr decreases of 9.8\% and 7.5\%, confirming its value for stable optimization. Figure~\ref{fig:ablation} illustrates these results.
The proposed framework obtained the best scores on all evaluation measures, outperforming the ablated versions on all the datasets consistently. The improvements are especially clear in MSVD and BDD-X, where temporal dependency modeling and interpreting intricate visual scenes are critical. This better performance is a result of the synergy among multiple important elements: ResNet50 for feature extraction of high-level spatial features from images, GPT-2 as a strong language model with the ability to produce fluent and coherent text sequences, multi-head attention for modeling fine-grained alignments between visual and linguistic modalities, and gradient accumulation with mixed-precision training for efficient and stable optimization. The use of GPT-2, specifically, adds to the model's capacity to generate contextually accurate and linguistically advanced descriptions. Collectively, these features make the proposed framework capable of generating high-quality, context-aware descriptions, with each module contributing significantly to the overall performance improvement.

\subsection{Results and Analysis:}
\begin{figure*}[!htbp]
    \centering
    \includegraphics[width=0.58\textwidth]{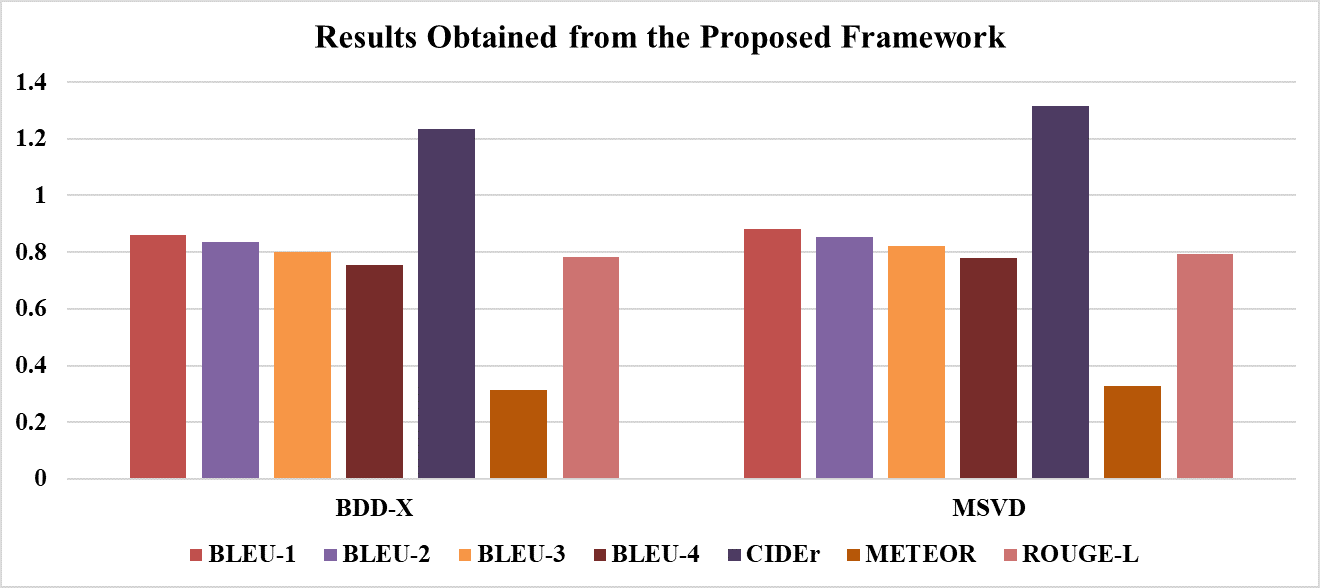}
    \caption{Graphical Representation of the Results Obtained}
    \label{fig:result}
\end{figure*}

To assess the efficacy of the suggested architecture, we carried out a comprehensive performance analysis on two benchmark datasets: MSVD and BDD-X. The suggested framework was tested against many model ablations, such as excluding gradient accumulation, using single-head attention, and removing ResNet-50. The evaluation's findings are compiled in Table \ref{results_analysis}.
\begin{table*}[!htbp] 
\centering 
\caption{Results and Analysis of the Proposed Framework} 
\label{results_analysis} 
\resizebox{\textwidth}{!}{ 
\begin{tabular}{|l|c|c|c|c|c|c|c|} 
\hline 
\textbf{Dataset} & \textbf{BLEU-1} & \textbf{BLEU-2} & \textbf{BLEU-3} & \textbf{BLEU-4} & \textbf{CIDEr} & \textbf{METEOR} & \textbf{ROUGE-L} \\ 
\hline
\textbf{BDD-X} & 0.860 & 0.835 & 0.800 & 0.755 & 1.235 & 0.312 & 0.782 \\
\hline
\textbf{MSVD} & 0.880 & 0.855 & 0.823 & 0.778 & 1.315 & 0.329 & 0.795 \\ 
\hline
\end{tabular}} 
\end{table*}
The outcomes shown in the table demonstrate how well our suggested methodology performs across datasets, consistently achieving top scores in all evaluation metrics. While high BLEU scores reflect fluency and grammatical accuracy, the superior performance in CIDEr and METEOR highlights strong semantic alignment and high-quality descriptions. With a BLEU-4 score of 0.778 and a CIDEr score of 1.315, the MSVD dataset showcases the model’s strength in producing diverse and precise captions for short video content. Similarly, the BDD-X dataset results—0.755 BLEU-4 and 1.235 CIDEr—illustrate the model’s robustness in capturing intricate driving scenes and generating contextually appropriate, descriptive narratives. Consistently high ROUGE-L scores (0.795 for MSVD and 0.782 for BDD-X) further validate the framework’s ability to generate outputs that closely align with human references. Notably, elevated CIDEr values emphasize the model’s effectiveness in capturing contextual and domain-specific nuances, while higher METEOR scores reflect improved alignment and paraphrasing capabilities. Figure~\ref{fig:result} demonstrates the graphical representation of the results obtained for the proposed framework.
In general, the findings demonstrate that the suggested approach greatly improves the ability of video-based descriptions, resulting in syntactically fluid, semantically rich, and contextually relevant descriptions.

\subsection{Comparison With State-of-the-Art Methods:}
\begin{figure*}[!htbp]
    \centering
    \includegraphics[width=0.80\textwidth]{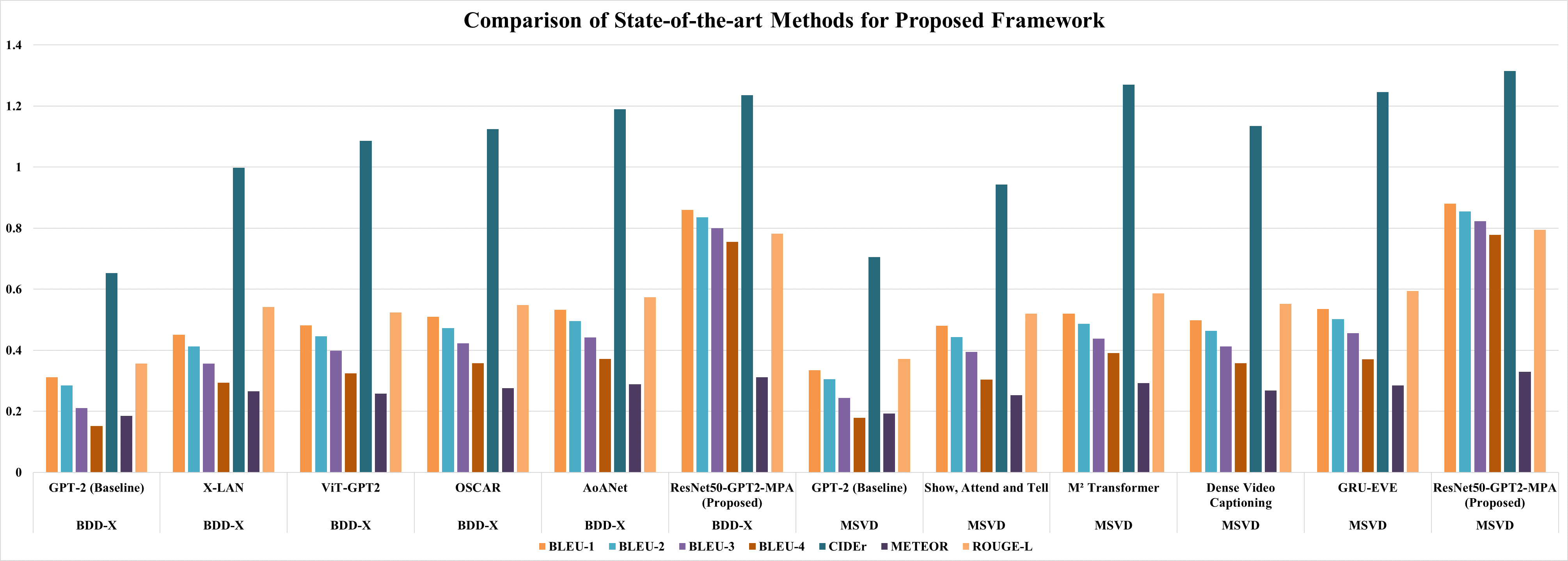}
    \caption{Graphical Representation of the SOTA Methods}
    \label{fig:sota}
\end{figure*}

We compare the performance of the suggested model with current state-of-the-art (SOTA) captioning methods on two benchmark datasets—BDD-X and MSVD—in order to assess its efficacy. A thorough evaluation employing several evaluation metrics, such as BLEU-1 to BLEU-4, CIDEr, METEOR, and ROUGE-L, is given by the results in Table~\ref{sota}. On both datasets, the proposed framework significantly outperforms earlier approaches, proving its capacity to produce more semantically precise and contextually rich descriptions for behaviours involving vehicles. Specifically, the suggested model leads in terms of BLEU-4 and CIDEr scores, demonstrating enhanced description coherence and relevance.
\begin{table*}[!htbp]
\centering
\caption{Comparison of State-of-the-Art Models for Vehicle Action Understanding}
\label{sota}
\resizebox{\textwidth}{!}{
\begin{tabular}{|l|l|c|c|c|c|c|c|c|}
\hline
\textbf{Dataset} & \textbf{Model} & \textbf{BLEU-1} & \textbf{BLEU-2} & \textbf{BLEU-3} & \textbf{BLEU-4} & \textbf{CIDEr} & \textbf{METEOR} & \textbf{ROUGE-L} \\ \hline

\multirow{6}{*}{BDD-X}  
& GPT-2 (Baseline) & 0.312 & 0.285 & 0.210 & 0.152 & 0.653 & 0.185 & 0.356 \\  
& X-LAN~\cite{pan2020x} & 0.451 & 0.412 & 0.356 & 0.293 & 0.998 & 0.265 & 0.541 \\  
& ViT-GPT2~\cite{vasireddy2023transformative} & 0.482 & 0.445 & 0.398 & 0.324 & 1.086 & 0.258 & 0.523 \\  
& OSCAR~\cite{li2020oscar} & 0.510 & 0.472 & 0.423 & 0.358 & 1.124 & 0.276 & 0.548 \\  
& AoANet~\cite{huang2019attention} & 0.532 & 0.495 & 0.442 & 0.372 & 1.189 & 0.289 & 0.573 \\  
& \textbf{ResNet50-GPT2 (Proposed)} & \textbf{0.860} & \textbf{0.835} & \textbf{0.800} & \textbf{0.755} & \textbf{1.235} & \textbf{0.312} & \textbf{0.782} \\ \hline  
\multirow{6}{*}{MSVD}  
& GPT-2 (Baseline) & 0.334 & 0.305 & 0.243 & 0.178 & 0.705 & 0.192 & 0.372 \\  
& Show, Attend and Tell~\cite{vinyals2015show} & 0.480 & 0.443 & 0.395 & 0.304 & 0.943 & 0.252 & 0.520 \\  
& M$^2$ Transformer~\cite{cornia2020meshed} & 0.520 & 0.486 & 0.438 & 0.391 & 1.270 & 0.292 & 0.586 \\  
& Dense Video Captioning~\cite{krishna2017dense} & 0.498 & 0.463 & 0.412 & 0.357 & 1.135 & 0.268 & 0.552 \\  
& GRU-EVE~\cite{aafaq2019spatio} & 0.535 & 0.502 & 0.456 & 0.370 & 1.246 & 0.285 & 0.594 \\  
& \textbf{ResNet50-GPT2 (Proposed)} & \textbf{0.880} & \textbf{0.855} & \textbf{0.823} & \textbf{0.778} & \textbf{1.315} & \textbf{0.329} & \textbf{0.795} \\ \hline  
\end{tabular}}
\end{table*}
The table's comparative analysis demonstrates how the BDD-X and MSVD datasets have advanced regarding vehicle action comprehension and general video-based description generation. Previous models like OSCAR~\cite{li2020oscar}, ViT-GPT2~\cite{vasireddy2023transformative}, X-LAN~\cite{pan2020x}, and AoANet~\cite{huang2019attention} showed consistent improvements in descriptive quality on the BDD-X dataset, with AoANet receiving the highest scores. However, with a BLEU-4 score of 0.755 and a CIDEr score of 1.235, the proposed model outperforms all of these approaches, significantly advancing over prior research. Similarly, on the MSVD dataset, transformer-based models such as M$^2$ Transformer~\cite{cornia2020meshed} and GRU-EVE~\cite{aafaq2019spatio} performed well, with GRU-EVE achieving a BLEU-4 score of 0.370 and CIDEr of 1.246. The proposed model demonstrates clear improvements, especially in BLEU-4 (0.778) and CIDEr (1.315), showcasing its ability to generate contextually rich and semantically aligned descriptions. Figure~\ref{fig:sota} depicts a graphical representation of SOTA methods.
Overall, the suggested approach continuously outperforms comparison on all datasets, setting new benchmarks and proving its ability to capture fine-grained contextual information for interpreting video-based actions.

\section{Conclusion and Future Direction}
\label{conclusion}
The proposed approach successfully combines ResNet50 for feature extraction with an improved GPT-2 model, allowing for the production of contextually aware, high-quality descriptions of video footage from the BDD-X and MSVD datasets. Through the use of transformer-based language modelling, the method guarantees descriptions that are consistent with ground-truth annotations and logical. In comparison to traditional approaches, the combination of ResNet50 for visual feature extraction and GPT-2 for text generation, as well as dataset augmentation, gradient accumulation, and mixed precision training, improves training efficiency and performance. Although transformer-based architectures are effective when used for large datasets, the study emphasizes that domain-specific fine-tuning is necessary to produce accurate and comprehensible descriptions. Future steps will involve implementing cross-modal attention mechanisms, sophisticated vision-based encoders, and optimized preprocessing methods to further improve model adaptability. The generalization and compatibility across other domains would be enhanced by expanding datasets to encompass a wider range of real-world circumstances. By enhancing interpretability, transparency, and decision-making clarity, this research advances explainable AI and increases the adaptability and dependability of automated video comprehension for a range of applications.

\section*{Acknowledgment}
The authors sincerely thank the Object-Oriented Programming Laboratory, Department of IT, DTU, Delhi, India, for providing the necessary resources to complete the research.

\section*{Statements and Declarations}

\textbf{Data Availability Statement:} The paper contains links to all datasets.\\
\textbf{Competing Interests:} The authors declare they have no competing financial interests.\\
\textbf{Conflict of Interest:} The authors declare no conflict of interest.

\bibliographystyle{sn-aps}
\bibliography{sn-article}
\end{document}